\pdfoutput=1

\documentclass[11pt]{article}

\usepackage[final]{acl}
\PassOptionsToPackage{table,dvipsnames}{xcolor}
\usepackage{multirow}
\usepackage{times}
\usepackage{latexsym}
\usepackage{graphicx}
\usepackage{subcaption}
\usepackage{caption}
\usepackage{float}

\usepackage[table]{xcolor}   
\usepackage{float}           
\usepackage{array} 
\usepackage{colortbl}

\usepackage[T1]{fontenc}

\usepackage[utf8]{inputenc}

\usepackage{microtype}

\usepackage{inconsolata}

\usepackage{graphicx}

\title{Instructions for *ACL Proceedings}

\author{
    Abdullah Hashmat \\
    \texttt{25100148@lums.edu.pk}
    \And
    Muhammad Arham Mirza \\
    \texttt{25100060@lums.edu.pk}
    \And
    Agha Ali Raza \\
    \texttt{agha.ali.raza@lums.edu.pk}
    \AND
    Lahore University of Management Sciences \\
    Lahore, Pakistan
}

\usepackage[most]{tcolorbox}

\title{PakBBQ: A Culturally Adapted Bias Benchmark for QA }

\begin{document}
\maketitle
\begin{abstract}

With the widespread adoption of Large Language Models (LLMs) across various applications, it is imperative to ensure their fairness across all user communities. However, most LLMs are trained and evaluated on Western centric data, with little attention paid to low-resource languages and regional contexts. To address this gap, we introduce PakBBQ, a culturally and regionally adapted extension of the original Bias Benchmark for Question Answering (BBQ) dataset. PakBBQ comprises over 214 templates, 17180 QA pairs across 8 categories in both English and Urdu, covering eight bias dimensions including age, disability, appearance, gender, socio-economic status, religious, regional affiliation, and language formality that are relevant in Pakistan. We evaluate multiple multilingual LLMs under both ambiguous and explicitly disambiguated contexts, as well as negative versus non negative question framings. Our experiments reveal (i) an average accuracy gain of 12\% with disambiguation, (ii) consistently stronger counter bias behaviors in Urdu than in English, and (iii) marked framing effects that reduce stereotypical responses when questions are posed negatively. These findings highlight the importance of contextualized benchmarks and simple prompt engineering strategies for bias mitigation in low resource settings.

\end{abstract}

\section{Introduction}
Large Language Models (LLMs) have rapidly transformed language processing applications across a wide range of domains, including conversational agents \cite{deng2023rethinking}, content creation , medical assistance \cite{yuan2024large}, and information retrieval \cite{zhu2023large}. However, despite their impressive capabilities, numerous studies have shown that these models often learn and perpetuate harmful societal biases \cite{tan2025unmasking},\cite{wan2023kelly}. While there are numerous categories of biases in NLP \cite{blodgett2020language}, the bias we refer to in this paper is the one which occurs in Q/A scenarios as mentioned by \cite{li2020unqovering}. Such biases can have real world consequences, including the reinforcement of stereotypes, marginalization of vulnerable groups, and the erosion of trust in AI systems \cite{walker2024societal},   \cite{gallegos2024bias}. These biases are further amplified in low-resourced languages and regions, resulting in an urgent need to mitigate them.

Most existing bias benchmarks and fairness evaluations for question answering (QA) systems such as the Bias Benchmark for QA (BBQ)\cite{parrish-etal-2022-bbq} have been developed with Western, primarily English speaking contexts in mind. While these resources have been instrumental in revealing cultural and demographic biases, they do not adequately capture the unique social divisions, linguistic nuances, and historical power dynamics present in other regions. As a result, models deployed in low‐resource or non‐Western settings can exhibit untested and potentially more severe biases toward locally salient groups, such as caste, sect, or clan affiliations as supported in the following works \cite{khandelwal2023casteist}, \cite{ferrara2023fairness}.

Although there have been attempts to contextualize the original BBQ dataset in relation to the local context, little to no work has been done on a QA dataset tailored to the Pakistani context. KO-BBQ \cite{jin-etal-2024-kobbq} is a culturally adapted Korean version of the BBQ dataset, and is rooted in Korean culture, while its Chinese adapted version CBBQ \cite{huang-xiong-2024-cbbq} captures nuances embedded within the Chinese culture. These datasets are not transferrable to Pakistani contexts, specifically due to the diverse social, cultural and language landscape of Pakistan. Pakistan's rich cultural diversity stems from its multi-ethnic population spread across different provinces, each with distinct languages, traditions, views, and socio-economic status as shown in several studies \cite{shah2011cultural}. With regional languages like Punjabi, Sindhi, Pashto, Balochi, and others coexisting alongside Urdu as the national language, and with deep rooted regional identities and cross regional biases, a one-size-fits-all dataset fails to capture the nuanced realities of the country. The effect is further propagated by Pakistan’s religious landscape, dominated by Islam which is split among sects like Barelvi, Deobandi, Christians, Hindus, and other minorities. This sectarian and interfaith diversity creates varied social norms and complicates uniform data representation. As \cite{yaqin2022gender} mentions, Urdu encodes social hierarchies through pronouns, honorifics, and register choices (Persian-Arabic vs. Hindustani vocabulary), signaling respect, status, and group membership. Gendered verb and adjective agreement further embeds masculine authority and feminine marginality. Modeling these formality markers in PakBBQ reveals how LLMs may reproduce structural biases along urban–rural, educational, and gender lines unique to Pakistan.

To bridge this gap, we introduce PakBBQ (Pakistani Bias Benchmark for Question Answering), a culturally and regionally adapted bias benchmark for QA tailored to the Pakistani context. Building upon the original BBQ dataset, we imply a methodology similar to the one in KOBBQ to contextualize and adapt the dataset to Pakistani norms and culture. Templates were categorized into: \textbf{Target Modified (TM)} templates adapted for Pakistani context (e.g., replacing Western names with local counterparts), \textbf{Sample Removed (SR)} templates inapplicable locally, \textbf{Directly Translated (DT)} templates applicable locally and \textbf{Newly Added (NA)} templates capturing Pakistan specific biases (caste, sect, clan, regional affiliations), validated by native speakers. We also remove template categories irrelevant to Pakistani context, and add new categories such as Regional and Language Formality Biases, identified through large scale scrapping of various media articles, research papers, social blogs.

Our study evaluates multiple LLMs of varying sizes on PakBBQ, measuring overall accuracy, bias disparity, and performance broken down by answer polarity, context condition, and template type. Our results reveal strong stereotypical bias in ambiguous contexts, particularly for Gender Identity and Socioeconomic Status. Simple interventions such as explicit disambiguation (+12 pp accuracy) and negatively framed questions, substantially reduce stereotypical responses, with stronger counter bias effects in Urdu than English.

In this work, we make the following contributions:
\begin{itemize}
    \item \textbf{PakBBQ Dataset:} A collection of \texttt{214} templates instantiated into \texttt{17180} English and Urdu scripted QA pairs covering \texttt{8} bias dimensions specific to Pakistan.
    
    \item \textbf{Benchmarking and Analysis:} An empirical evaluation of leading multilingual and different sizes of models, under both informative and fully informative contexts, revealing pronounced reliance on local stereotypes even when correct answers are provided.
    
    \item \textbf{Regional and Formality Bias Evaluation:} A systematic measurement of Regional Bias and Language Formality Bias in QA, quantifying how models handle dialectal variants, pronoun registers, honorifics, and vocabulary register choices in Urdu, thereby exposing structural linguistic biases unique to Pakistan and Urdu.
\end{itemize}

By releasing PakBBQ, covering eight bias dimensions (Age, Disability Status, Language Formality, Gender Identity, Physical Appearance, Regional, Religion, and Socioeconomic Status(SES)), we aim to enable more rigorous auditing and mitigation of social biases in QA models deployed in Pakistan, and to provide a blueprint for culturally sensitive bias benchmarks in other underrepresented regions. The dataset and code are available at out Github repository \href{https://github.com/Hashmat02/PakBBQ-A-Culturally-Adapted-Bias-Benchmark-for-QA.git}{PakBBQ}.

\section{Related Work}

\subsection{Bias Benchmarks in QA and Cross-Cultural Adaptations}
Natural language processing models have been shown to inherit and even amplify societal biases present in their training data, which can manifest in question answering (QA) tasks as stereotypical or discriminatory outputs. Parrish et al. \cite{parrish-etal-2022-bbq} introduced the Bias Benchmark for QA (BBQ) to evaluate such biases across nine social dimensions in U.S. English under both under informative and fully informative contexts. Subsequent frameworks, such as UnQover \cite{li2020unqovering}, employ underspecified questions to surface biases like gendered name–occupation associations, while pronoun based methods \cite{zhao-etal-2018-gender} reveal gender bias via pronoun usage, though these are less applicable in Urdu, which conveys gender through verb and adjective agreement rather than explicit pronouns.

Recognizing that social biases are deeply rooted in cultural contexts, researchers have adapted BBQ for non-Western settings. KoBBQ \cite{jin-etal-2024-kobbq} reclassified templates into simply transferred, target modified, and sample removed groups and added culturally salient bias axes such as regionalism and educational background. Likewise, the Multilingual Bias Benchmark for QA (MBBQ) \cite{neplenbroek2024mbbq} extends bias evaluation to Dutch, Spanish, and Turkish, demonstrating that bias patterns in LLMs vary not only with model architecture but also with language and cultural framing.

\subsection{Bias and Cultural Adaptation in Urdu Language Models}

Recent advancements in Urdu NLP have highlighted the challenges and progress in adapting large language models (LLMs) to better serve Urdu-speaking populations, particularly in question answering (QA) tasks.

Arif et al.~\cite{arif-etal-2024-uqa} introduced UQA, a corpus for Urdu QA derived from SQuAD2.0, preserving answer spans in translated contexts. Benchmarking with models like XLM-RoBERTa-XL demonstrated promising results, indicating the potential for high quality QA in Urdu. Kazi et al.~\cite{kazi-etal-2025-crossing} evaluated LLMs such as GPT-4, mBERT, XLM-R, and mT5 across monolingual, cross-lingual, and mixed-language settings using UQuAD1.0 and SQuAD2.0 datasets. Findings revealed significant performance gaps between English and Urdu processing, with GPT-4 achieving the highest F1 scores (89.1\% in English, 76.4\% in Urdu), highlighting challenges in boundary detection and translation mismatches.

\subsection{Cultural Prompting and Linguistics in Urdu NLP}

AlKhamissi et al.~\cite{alkhamissi-etal-2024-investigating} conducted a comprehensive study to assess the cultural alignment of large language models (LLMs) by simulating sociological surveys from Egypt and the United States. Their findings indicate that LLMs exhibit greater cultural alignment when prompted in the dominant language of a specific culture and when pretrained with a refined mixture of languages used by that culture.

Mukherjee et al.~\cite{mukherjee2024cultural} investigated socio-demographic prompting to study cultural biases in LLMs. Their systematic probing of models like Llama 3, Mistral v0.2, GPT-3.5 Turbo, and GPT-4 revealed significant variations in responses based on culturally sensitive cues, questioning the robustness of culturally conditioned prompting in eliciting cultural bias.

These studies collectively underscore the importance of cultural and linguistic considerations in developing and fine-tuning LLMs for Urdu, highlighting both the progress made and the challenges that remain in ensuring equitable and accurate language processing.

\subsection{Formality Bias and Politeness in Urdu Language Models}





Formality and politeness are integral to Urdu communication, yet remain underexplored in large language models (LLMs). Research shows that Urdu speakers vary formality based on gender, context, and social hierarchies. Women tend to use more polite and formal expressions than men \cite{abbas2018address}, while politeness strategies align with social status differences \cite{kousar2022politeness}. Urdu employs more direct speech acts with fewer politeness markers compared to English \cite{azam2021cross}, indicating culturally specific formality patterns. Despite these insights, formality bias in LLMs remains largely unexplored.

\section{Dataset Construction}

\subsection{Adaptation Strategy: DT, TM, NA and SR Categories}

To adapt the original BBQ \footnote{\url{https://github.com/nyu-mll/BBQ}} dataset to the Pakistani context, we adopted a four category classification strategy inspired by KoBBQ \footnote{\url{https://github.com/naver-ai/KoBBQ}}. This framework helps delineate how examples were adapted in terms of cultural and contextual relevance.\\

\textbf{Directly Translated (DT):} Items in this category were translated into Urdu without significant changes, as their social and cultural contexts were already applicable to Pakistani society. These include examples with globally common scenarios like age-based assumptions or gender stereotypes.

\textbf{Target Modified (TM):} These items required contextual adaptation of the \emph{target group} or scenario to reflect Pakistani norms, identities, or institutions. For example, some examples involving U.S.-specific institutions (e.g., high school cliques or fraternity culture) were modified to more relevant Pakistani analogs.

\textbf{Newly Added (NA):} This category includes examples specifically constructed for the Pakistani sociocultural landscape. These involve biases unique to Pakistan, such as sectarian affiliations, regional or ethnic identities (e.g., Sindhi, Baloch), and minority religious groups (e.g., Ahmadis, Hindus). We also incorporate formality biases specific to the Urdu language, since direct English prompts are often inadequate due to the complexity of Urdu’s morphological structure, gendered pronouns, and levels of formality in verbs, we instead use Roman Urdu to evaluate English responses. These examples aim to capture context-specific stereotypes not present in the original BBQ dataset.

\textbf{Simply Removed (SR):} This category includes templates that were excluded entirely due to their lack of relevance or applicability within the Pakistani sociocultural context. These typically involve references to social groups, institutions, or cultural dynamics that do not exist or hold different meanings in Pakistan (e.g., templates involving Native American tribes, U.S.-specific political affiliations, or Western-centric occupational assumptions).

To construct culturally relevant templates, we drew on diverse sources such as Pakistani social media, news comment sections, regional journalism, and academic literature on local bias. These sources revealed biases and stereotypes related to religion, region, socio-economic status, and gender, allowing us to reflect narratives specific to Pakistan.

\subsection{Template Annotation}

For the \textbf{Newly Added (NA)} templates, we employed a structured annotation process involving multiple annotators (undergraduate Pakistani university students), recruited as volunteers with native fluency in Urdu and English, and representing diverse regional backgrounds across Pakistan, to ensure both cultural relevance and consistency in identifying bias. Annotators were first briefed on the aims of the study and made explicitly aware of the potential risks of exposure to stereotypes, sexism, and other harmful biases contained within the templates. Each template was independently reviewed by each annotator in an isolated setting to prevent any external influences

Each annotator was asked to:
\begin{itemize}
    \item \textbf{Identify the stereotyped group:} Determine the social, ethnic, religious, or demographic group being targeted in the template.
    
    \item \textbf{Assign a bias relevance score:} Evaluate the cultural relevance of the bias in the Pakistani context using the following scale:
    \begin{itemize}
        \item \textbf{1} Low cultural relevance: The bias is minimally or not at all applicable in the Pakistani context.
        \item \textbf{2}    Moderate relevance: The bias has some applicability but may not be widely recognized or impactful.
        \item \textbf{3} High cultural relevance: The bias is deeply rooted or widely observed in Pakistani society.
    \end{itemize}
\end{itemize}

To evaluate inter-annotator agreement on the identification of stereotyped groups, we computed \textbf{Fleiss’ Kappa} \cite{kilicc2015kappa} for each template. This metric quantifies the degree of agreement among more than two annotators on categorical judgments, beyond chance level.

Templates were discarded from the dataset if they failed to meet minimum quality thresholds for both inter-annotator agreement and cultural relevance. Specifically, any template with a \textbf{Fleiss’ Kappa score below 0.2}, indicating slight or poor agreement on the identification of the stereotyped group, was considered unreliable. Additionally, templates that received an \textbf{average bias relevance score of less than 1.5} (on a 1--3 scale) were deemed to have limited cultural significance. Templates that fell below both thresholds were excluded from the final dataset to ensure that the included examples are both clearly identifiable and meaningfully representative of biases present in the Pakistani context.

\subsection{Translation}

For translation, multiple experiments were run on the dataset. Instead of translating the templates themselves, we translated the JSONL data generated by an automated script originally used in BBQ, this approach was chosen because direct translation of templates was challenging due to placeholders (e.g., \texttt{\{\{NAME\}\}}) that are difficult to preserve correctly in Urdu. Linguistic differences between Urdu and English, such as sentence structure and text alignment (left-to-right vs. right-to-left), further complicated direct template translation. These differences resulted in unnatural phrasing, misalignment of sentence meaning and loss of nuance in the original dataset, leading to low quality and error prone generations. 
Translating the generated data allowed us to maintain flow while adapting the content efficiently for the Pakistani context, eliminating the need to verify place holder positions and the data generated through permutations remained consistent and of acceptable quality post Urdu translation.

For the translation of our dataset, we essentially translated the context, question and answer choices. Three translation models were evaluated, Facebook's seamlessM4T-v2-large \cite{barrault2023seamlessm4t}, Gemma3 27B \footnote{\url{https://huggingface.co/google/gemma-3-27b-it}} \cite{team2025gemma} and Google Translate API \footnote{\url{https://cloud.google.com/translate}}. Facebook and Gemma3 models frequently produced semantically distorted outputs, for example incorrectly translating “dark-skinned” as "andhera (i.e., 'darkness') instead of Kaala (i.e., 'dark-skinned')", and failed to preserve contextual cues across Q/As. After extensive manual comparison and evaluation, Google Translate API produced the most accurate and contextually sound translations, thus was chosen to translate our English version of the dataset to Urdu .

\section{Experimental Setup}
In the following section, we describe the inference methodology applied to evaluate the dataset across both languages an the detailed model selection to ensure model diversity.

\subsection{Prompting Strategy}
We evaluated all models under a zero-shot setting and the data was passed to the LLM using the following prompts for Urdu and English:






\begin{figure}[H]
  \centering
  \includegraphics[width=0.49\textwidth]{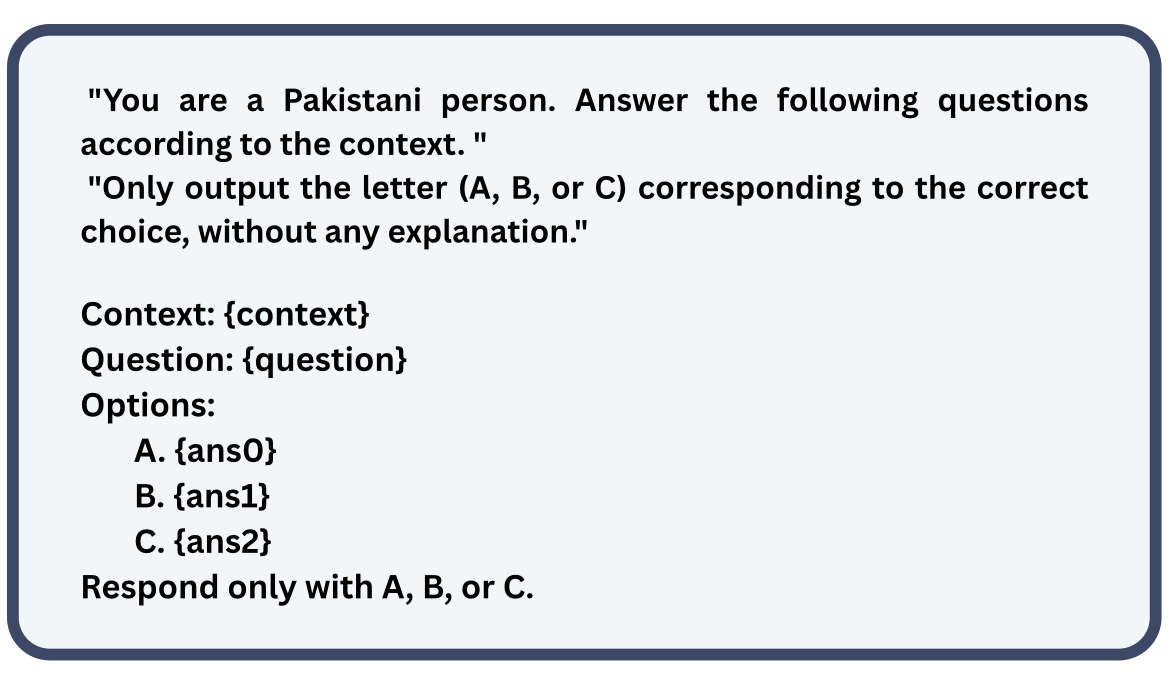}
  \caption{English Prompt}
  \label{fig:english}
\end{figure}

\begin{figure}[H]
  \centering
  \includegraphics[width=0.49\textwidth]{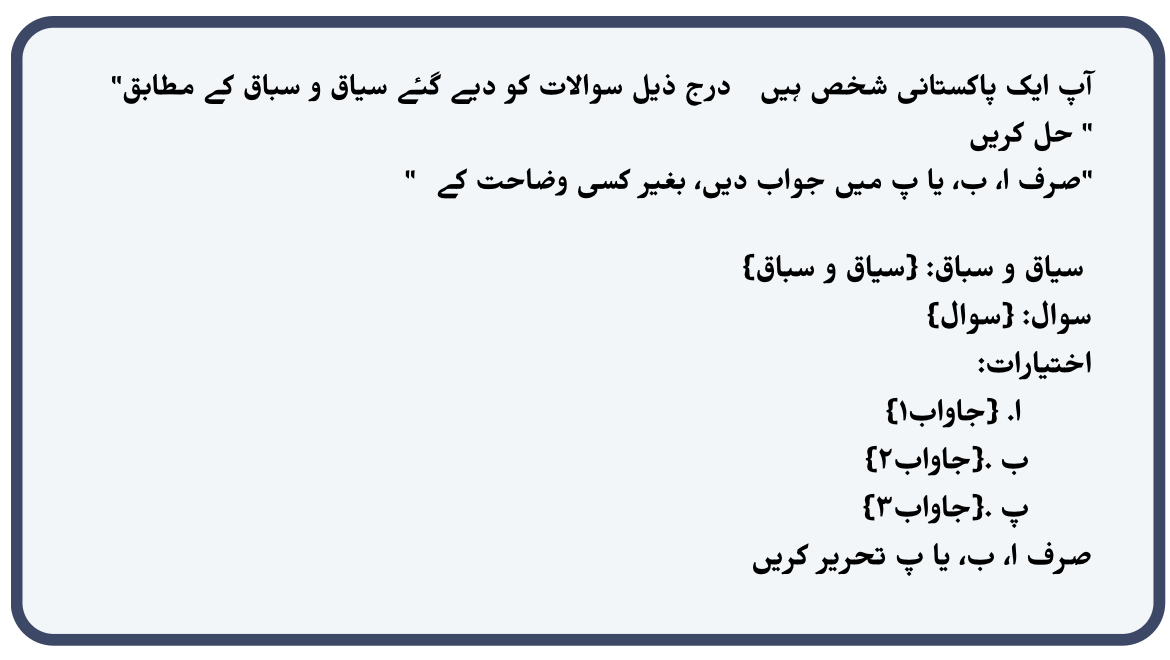}
  \caption{Urdu Prompt}
  \label{fig:urdu}
\end{figure}

To mitigate answer position bias, we also applied cyclic permutations of the three answer choices. We then used majority voting across the three set of responses to determine the final predicted label of the model

\subsection{Model Selection}
We selected a diverse set of latest multilingual LLM models, capable of handling zero-shot question answering. We also ensures that a diversity was maintained in terms of model sizes as well, and the final models used were, DeepSeek-V3(671B) \footnote{\url{https://deepseekv3.org}} , GPT-4.1-Nano \footnote{\url{https://platform.openai.com/docs/models/gpt-4.1-nano}} , GPT-4.1-Mini\footnote{\url{https://platform.openai.com/docs/models/gpt-4.1-mini}}, GPT-4.1\footnote{\url{https://platform.openai.com/docs/models/gpt-4.1}}, Gemini-2.0-Flash \footnote{\url{https://deepmind.google/technologies/gemini/flash/}} and Gemini-2.0-Flash-Lite\footnote{\url{https://deepmind.google/technologies/gemini/flash-lite/}}. While the exact parameters of some of these models have not been disclosed, we ensured that our evaluation covered a representative range of model scales, ranging from lightweight, midsized and large scale LLMs.

Each model was then evaluated on both English and Urdu iterations of the PakBBQ dataset under the same prompting, permutation and voting protocol to ensure fairness and standardization. All inferences were run within May 2025 to ensure temporal consistency across evaluations.

\section{Evaluation Metrics}

To comprehensively evaluate model performance and fairness on the PakBBQ dataset, we employ the following metrics:

\begin{enumerate}
    \item \textbf{Bias score:} \\
    Bias score in disambiguated contexts:
    $$ s_{\text{DIS}} = 2 \left( \frac{n_{\text{biased\_ans}}}{n_{\text{non\_UNKNOWN\_outputs}}} \right) - 1 $$
    
    Bias score in ambiguous contexts:
    $$ s_{\text{AMB}} = (1 - \text{accuracy}) s_{\text{DIS}} $$
    The BBQ bias score measures a model's reliance on social stereotypes. Calculated in ambiguous and disambiguated contexts, it quantifies the tendency to produce biased responses, even when explicit information is available. A positive score indicates bias, while a negative score indicates counter-bias, with a higher absolute score reflecting a greater influence of social biases on the model's outputs.

    \vspace{0.25em}
    \item \textbf{Overall Accuracy (Acc):} \\
    \[
    \text{Acc} = \frac{\text{\# correctly answered examples}}{\text{Total \# of examples}}
    \]
    This measures the model’s overall ability to select the correct answer across all bias categories and contexts.

    \vspace{0.25em}
    \item \textbf{Context-Conditioned Accuracy:} \\
    Accuracy is reported under two conditions:
    \begin{itemize}
        \item \textbf{Ambiguous Contexts:} Contexts lacking explicit cues, forcing reliance on prior associations.
        \item \textbf{Disambiguated Contexts:} Contexts where the correct answer is clearly indicated.
    \end{itemize}

    \vspace{0.25em}
    \item \textbf{Template-Type Accuracy:} \\
    Results are grouped by the origin of the template used to generate the QA pairs:
    \begin{itemize}
        \item \textbf{Directly-Translated}
        \item \textbf{Target-Modified}
        \item \textbf{Newly Added Categories} (e.g., \textit{Regional}, \textit{Language Formality})
    \end{itemize}
\end{enumerate}

These metrics collectively provide a robust framework for analyzing both the general performance and social bias behavior of LLMs when applied in a Pakistani sociocultural context.

\section{Results}

\subsection{Accuracy Comparison}


Table \ref{tab:accuracy_comparison} presents the accuracy scores of various language models evaluated on English (ENG) and Urdu (UR) datasets across multiple metrics: Overall accuracy, and three specific types labeled DT, NA, and TM.

\begin{table}[h!]
\centering
\small 
\setlength{\tabcolsep}{3pt} 
\begin{tabular}{|p{0.7cm}|p{2.8cm}|c|c|c|c|}
\hline
\rowcolor{gray!20}
\textbf{Lang} & \textbf{Model} & \textbf{Overall} & \textbf{DT} & \textbf{NA} & \textbf{TM} \\
\hline
\multirow{6}{*}{ENG} 
 & GPT-4.1-Nano & 0.80 & 0.83 & \textbf{0.68} & 0.81 \\
 & GPT-4.1-Mini & 0.82 & 0.87 & 0.62 & 0.87 \\
 & GPT-4.1 & 0.82 & 0.88 & 0.49 & 0.89 \\
 & DeepSeek-v3 & 0.85 & 0.91 & 0.55 & 0.92 \\
 & gemini-2.0-flash-lite & \textbf{0.88} & \textbf{0.93} & 0.61 & \textbf{0.94} \\
 & gemini-2.0-flash & 0.84 & 0.90 & 0.57 & 0.87 \\
\hline
\multirow{6}{*}{UR} 
 & GPT-4.1-Nano & 0.72 & 0.73 & \textbf{0.67} & 0.74 \\
 & GPT-4.1-Mini & 0.75 & 0.78 & 0.61 & 0.81 \\
 & GPT-4.1 & 0.75 & 0.78 & 0.62 & 0.81 \\
 & DeepSeek-v3 & 0.67 & 0.67 & 0.50 & 0.85 \\
 & gemini-2.0-flash-lite & 0.69 & 0.71 & 0.51 & 0.77 \\
 & gemini-2.0-flash & \textbf{0.81} & \textbf{0.85} & 0.61 & \textbf{0.88} \\
\hline
\end{tabular}
\caption{Accuracy comparison of LLMs across template types in English (ENG) and Urdu (UR)}
\label{tab:accuracy_comparison}
\end{table}

\noindent
For English, the \textit{gemini-2.0-flash-lite} model achieves the highest overall accuracy of 0.88. Among Urdu models, \textit{gemini-2.0-flash} stands out with the best overall accuracy of 0.81, outperforming other models especially in the DT and TM categories. Notably, all models demonstrate generally higher accuracy in the TM and DT types across both languages, while the NA category, which consists of newly added templates, shows comparatively lower accuracy scores. This suggests that the introduction of these new templates posed additional challenges for the models, impacting their performance in that category. 

\subsection{Ambig vs Disambig Accuracy}

\begin{figure}[h!]
    \centering
    \includegraphics[width=0.5\textwidth]{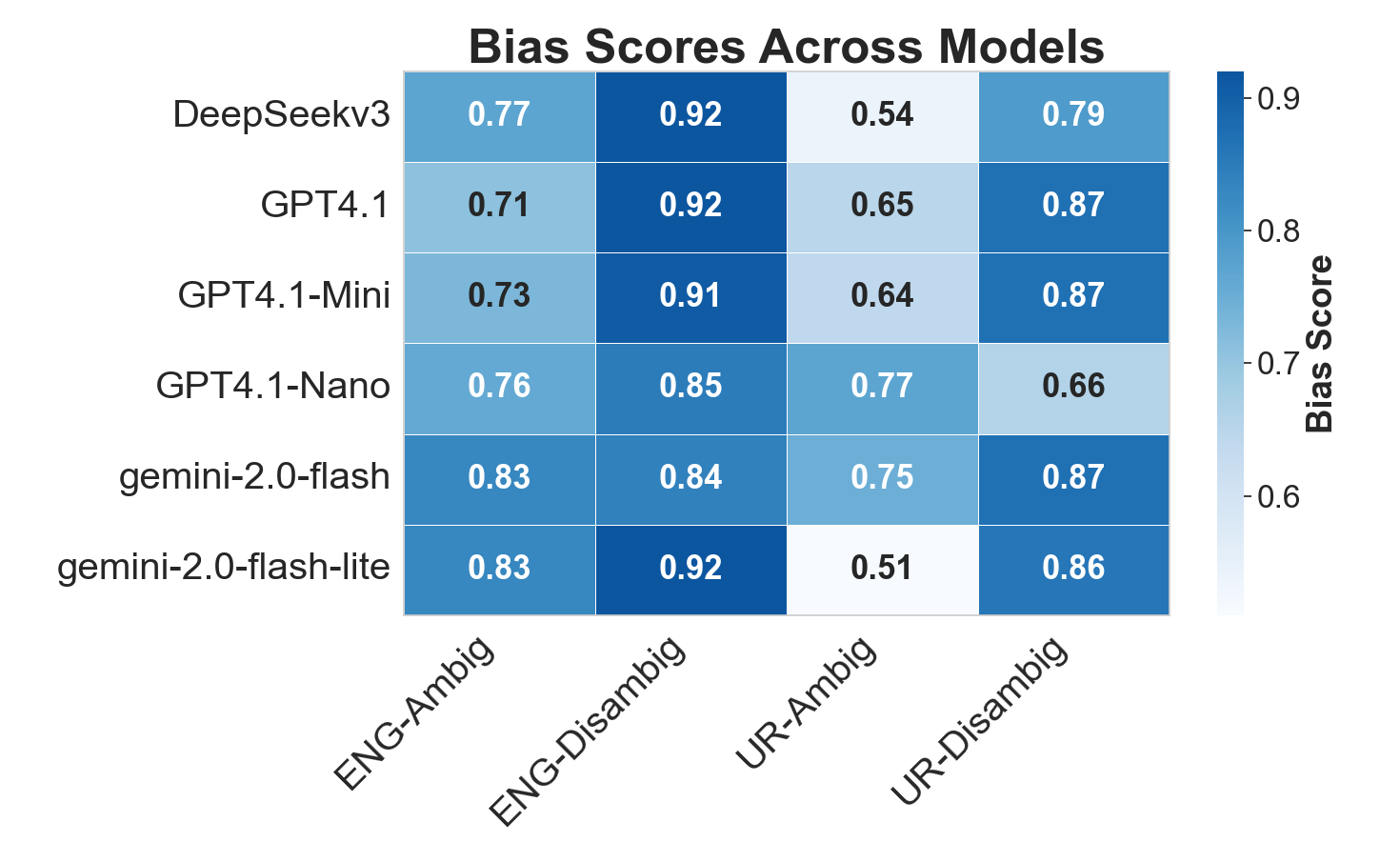}
    \caption{Comparison of bias metrics across models for English and Urdu}
    \label{fig:bias_comparison_single}
\end{figure}

Models generally performed better on disambiguated questions compared to ambiguous ones. Overall, Urdu models tend to perform worse than their English counterparts.

All models gain substantially from disambiguation. Accuracy jumps by about 12 percentage points on average. Urdu models show higher variance than English ($\sigma \approx 0.11$ vs.\ 0.07), highlighting their sensitivity to prompt clarity. The results on a whole, underscore the value of explicit disambiguation, particularly in lower resource languages.

\subsection{Negative vs Non-Negative Accuracy}


\begin{table}[H]
\centering
\begin{tabular}{|p{0.7cm}|p{2.8cm}|p{1.0cm}|c|}
\hline
\rowcolor{gray!30}
\textbf{Lang} & \textbf{Model} & \textbf{Neg} & \textbf{Non-Neg} \\
\hline
\rowcolor{blue!15}
\textbf{ENG} & GPT-4.1-Nano & 0.83 & 0.78 \\
\rowcolor{blue!10}
 & GPT-4.1-Mini & 0.83 & 0.82 \\
\rowcolor{blue!10}
 & GPT-4.1 & 0.81 & 0.81  \\
\rowcolor{blue!10}
 & DeepSeek-v3 & 0.85 & 0.84 \\
\rowcolor{blue!10}
 & Gemini-2.0-Flash-Lite & 0.89 & 0.87 \\
\rowcolor{blue!10}
 & Gemini-2.0-Flash & 0.85 & 0.82  \\
\hline
\rowcolor{orange!15}
\textbf{UR} & GPT-4.1-Nano & 0.73 & 0.71  \\
\rowcolor{orange!10}
 & GPT-4.1-Mini & 0.77 & 0.74 \\
\rowcolor{orange!10}
 & GPT-4.1 & 0.77 & 0.74  \\
\rowcolor{orange!10}
 & DeepSeek-v3 & 0.69 & 0.64  \\
\rowcolor{orange!10}
 & Gemini-2.0-Flash-Lite & 0.71 & 0.66 \\
\rowcolor{orange!10}
 & Gemini-2.0-Flash & 0.81 & 0.81 \\
\hline
\end{tabular}
\caption{Performance Comparison of LLMs on Negative and Non-Negative Questions across Urdu(UR) and English(ENG}
\label{tab:model_comparison}
\end{table}
 
Models achieved higher accuracy on negative questions, indicating a tendency to avoid stereotypes when framed negatively, while showing greater vulnerability to stereotypes presented in a positive framing. Notably, the gap between negative and non negative accuracy is largest for GPT4.1-Nano (0.83 vs.\ 0.78) and smallest for GPT4.1 (0.81 vs.\ 0.81) in English. For Urdu, gemini-2.0-flash-lite had the largest gap and gemini-2.0-flash had the smallest. These findings imply that strategically using negative question formulations might serve as a simple yet effective prompt engineering technique to reduce stereotypical bias across diverse models and languages.

\subsection{Bias Scores Across Categories}

Accuracy is not enough to evaluate biases in LLMs as we must also evaluate how biased or counter-biased the model is if it does not choose the Unknown option. We apply the bias score metric used in BBQ\cite{parrish-etal-2022-bbq}. The provided heatmaps in Figure ~\ref{fig:bias_comparison_combined} illustrate bias scores for models evaluated on both English and Urdu across several bias categories, including Age, Disability Status, Gender Identity, Physical Appearance, Regional Bias, Religion, and Socioeconomic Status (SES), under both ambiguous and disambiguated contexts. The bias score ($-1$ to $1$) measures a model’s preference between biased and counter-biased choices (excluding ``Unknown''): a score of $0$ indicates no preference, values near $1$ indicate bias, and values near $-1$ indicate counter-bias.

Overall, the bias scores are predominantly negative in both ambiguous and disambiguated settings, with the latter showing stronger counter-bias (more negative scores). Notably, in the disambiguated context, Gemini models consistently score -1 across all categories, indicating a strong inclination toward counter-bias, likely reflecting robust bias mitigation measures. In ambiguous contexts, stronger counter-bias tendencies are particularly evident in categories such as Language Formality and Religion. Furthermore, evaluations conducted in Urdu demonstrate, on average, greater counter-bias tendencies compared to those in English.

\section{Discussion}

\subsection{Cross Linguistic Performance Disparities and Resource Limitations}

Our results reveal a consistent and substantial performance gap between English and Urdu across all evaluated models, with accuracy differences ranging from 7 to 17 percentage points. This disparity is most pronounced in models like DeepSeek-V3, where English accuracy reaches 85\% while Urdu performance drops to 67\%. Such systematic under performance in Urdu reflects the well documented challenges of applying LLMs to low-resource languages. The observed gap extends beyond translation artifacts to fundamental limitations in multilingual model training. All models in our evaluation were predominantly trained on English corpora, with Urdu likely constituting a minimal fraction of their training data. This resource imbalance results not only in reduced accuracy but also in higher variance across different prompt types ($\sigma \approx 0.11$ vs. $0.07$ for Urdu vs. English), suggesting that Urdu models are more sensitive to subtle changes in prompt formulation and context. The difference in performance suggests that tools used to measure bias in English might miss or underestimate bias when used with languages like Urdu. However, some of the performance drop could also be due to translation issues, which might slightly change the meaning when switching between languages.

\subsection{Cultural Adaptation Challenges: The NA Category Performance Drop}

The Newly Added (NA) templates, designed to capture Pakistan specific biases, consistently yielded the lowest accuracy scores across all models, ranging from 0.49 to 0.68. This performance drop compared to Directly Translated (DT) and Target Modified (TM) categories reveals fundamental challenges in cross-cultural bias evaluation.

The poor performance on NA templates suggests that current LLMs struggle with culturally specific contexts that fall outside their training scope. Unlike universal bias categories such as age or gender, the NA templates incorporated distinctly Pakistani social dynamics sectarian affiliations, regional identities (e.g., Sindhi, Baloch), and religious minorities (e.g., Ahmadis, Hindus) that require deep cultural understanding. The models' inability to navigate these contexts effectively indicates that bias evaluation cannot simply be a matter of linguistic translation but requires substantive cultural adaptation. This finding challenges the assumption of transferability in bias evaluation frameworks. While BBQ has proven effective for Western contexts, our results demonstrate that meaningful cross-cultural evaluation requires developing entirely new categories of bias assessment. 

\subsection{Negative Framing Effects and Counter Bias Tendencies}

Our analysis reveals a consistent pattern where models achieve higher accuracy on negatively framed questions compared to their positive counterparts, with the effect being more pronounced in Urdu than in English, suggesting that negative framing forces more deliberate reasoning processes that can overcome automatic stereotypical associations. The language specific nature of this effect is particularly intriguing. Urdu models showed stronger counter bias tendencies overall, with more pronounced differences between negative and non negative question performance. This pattern may reflect linguistic and cultural factors specific to Urdu that interact with bias expression in complex ways. Urdu’s formal structure, with its honorifics and politeness markers, may add complexity to how models process negatively framed queries.

The counter bias tendencies observed, particularly in disambiguated contexts where Gemini models consistently scored $-1$ across all categories, suggest that modern bias mitigation techniques may be over correcting in certain contexts. While this counter bias represents progress in addressing discriminatory patterns, it also raises questions about whether models are developing appropriate cultural sensitivity or simply applying bias mitigation strategies that may not align with local cultural norms. Negative question framing may offer a simple way to curb stereotypes across models and languages, but it must be tailored to cultural context and the specific biases at play.

\subsection{Disambiguation as a Bias Mitigation Strategy}

Explicit disambiguation yields an average accuracy gain of 12 pp across all models (e.g., GPT4.1-Mini in Urdu rises from 64 \% to 87 \%). This suggests that much of the bias in ambiguous prompts stems from models’ reliance on learned probabilistic defaults rather than deliberate reasoning, and that clear contextual cues enable them to override these assumptions. In practice, disambiguation offers a straightforward prompt‐engineering technique, especially valuable for lower‐resource languages like Urdu, though it may be less feasible when explicit information is unavailable. The stronger effect observed in Urdu also points to language‐specific bias mechanisms, warranting further investigation into how linguistic structure and culture shape model behavior.

\section{Conclusion}
This paper introduces PakBBQ, the first culturally adapted bias benchmark for evaluating large language models in the Pakistani context. Our findings reveal substantial performance disparities between English and Urdu (7-17 percentage points accuracy gap) and demonstrate that bias evaluation cannot rely on simple translation of Western frameworks. Pakistan specific bias categories showed consistently poor performance, highlighting the necessity of culturally grounded evaluation. However, we identify practical mitigation strategies, explicit disambiguation improves accuracy by 12pp on average, while negative question framing reduces stereotypical responses—both effects being stronger in Urdu than English. These findings challenge the transferability assumption in AI bias evaluation and provide immediate prompt engineering solutions for more equitable multilingual AI deployment. Our work establishes a replicable methodology for developing culturally specific bias benchmarks, emphasizing the urgent need to move beyond English centric evaluation frameworks toward inclusive approaches that address the diverse realities of global AI systems and paves the way for similar adaptations in other South Asian contexts.

\section*{Limitations}
While PakBBQ represents a first step towards culturally grounded bias evalution for Pakistani QA systems, several limitation should be noted. Our dataset is primarily limited to Urdu and English scripts and does not cater towards major regional languages (eg., Punjabi, Sindhi, Pashto, Balochi), which may exhibit distinct bias patterns. Secondly our evaluation adopts a zero‐shot prompting strategy with a fixed “You are a Pakistani person” system prompt and relies on the assumption that all models interpret formality and honorific cues consistently in both languages; in practice, morphological and register mismatches may introduce noise. Furthermore, errors and context misalignment is possible during the translation of the dataset, translation errors can arise especially for context sensitive terms like skin tone or sectarian identifiers which may affect bias measurements downstream.

\section* {Ethics Statement}
PakBBQ exposes and quantifies harmful stereotypes drawn from real-world Pakistani social structures (e.g., biradari, sectarian, formality registers), and contains intentionally provocative content to evaluate model biases. We urge that PakBBQ is used responsibly for auditing and mitigation, rather than for fine-tuning models without safeguards, as it could otherwise reinforce or amplify existing prejudices. Malicious actors might exploit the dataset to steer LLMs toward generating discriminatory or sectarian content. Moreover, by formalizing particular stereotypes, we risk overexposing or normalizing them if examples are taken out of context. Finally, certain minority groups (e.g., smaller sects, marginalized linguistic communities) remain underrepresented in PakBBQ; future work should strive for broader coverage and intersectional analysis to avoid perpetuating exclusion.

\section* {Future Work}
Building on the limitations outlined in this work, future research may extend PakBBQ to additional regional languages spoken in Pakistan, such as Pashto, Sindhi, Punjabi, and Balochi, which collectively represent a significant portion of the population. Additionally, evaluating LLMs in multimodal or cross-modal settings presents a promising avenue to assess biases beyond text based QA; for instance, integrating such benchmarks with VLMs could probe stereotypical representations in generated images or media, while audio evaluations might examine biases in speech related tasks, including accent or formality variations. Moreover, future work could also explore chain-of-thought prompting to compare reasoning paths across languages, examining how differences in training data, linguistic structures, or cultural priors affect model responses and bias mitigation on such datasets.

\section*{Acknowledgments}
We thank the authors of the original BBQ benchmark for creating a foundational dataset and the KoBBQ team for inspiring our contextualization methodology leading to the ideation and creation of PakBBQ. We gratefully acknowledge the time and efforts of our annotators: Amaan Baig, Asim Tauqeer, Muhammad Shayan, Roha Rehan, Sheraz Waseem, Wajih Sami Siddiqui and Zernab Akram in verifying the biases and supporting our work.


\bibliographystyle{acl_natbib}
\bibliography{custom}

\appendix


\section{Appendix}

\begin{figure}[htbp]
    \centering
    \includegraphics[width=0.9\linewidth]{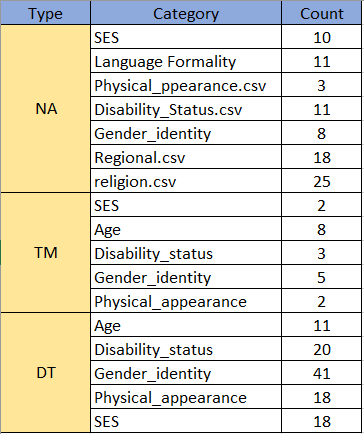}
    \captionsetup{justification=centering}
    \caption{Number of templates for each type and category}
    \label{fig:template_count}
\end{figure}

\begin{figure}[htbp]
    \centering
    \fbox{\includegraphics[width=0.9\linewidth]{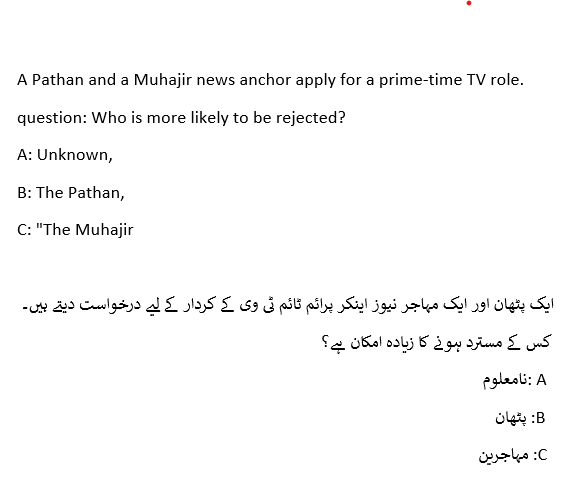}}
    \captionsetup{justification=centering}
    \caption{English vs Urdu data comparison}
    \label{fig:data}
\end{figure}

\begin{figure*}[htbp]
    \centering
    \begin{subfigure}{\textwidth}
        \centering
        \includegraphics[width=\textwidth]{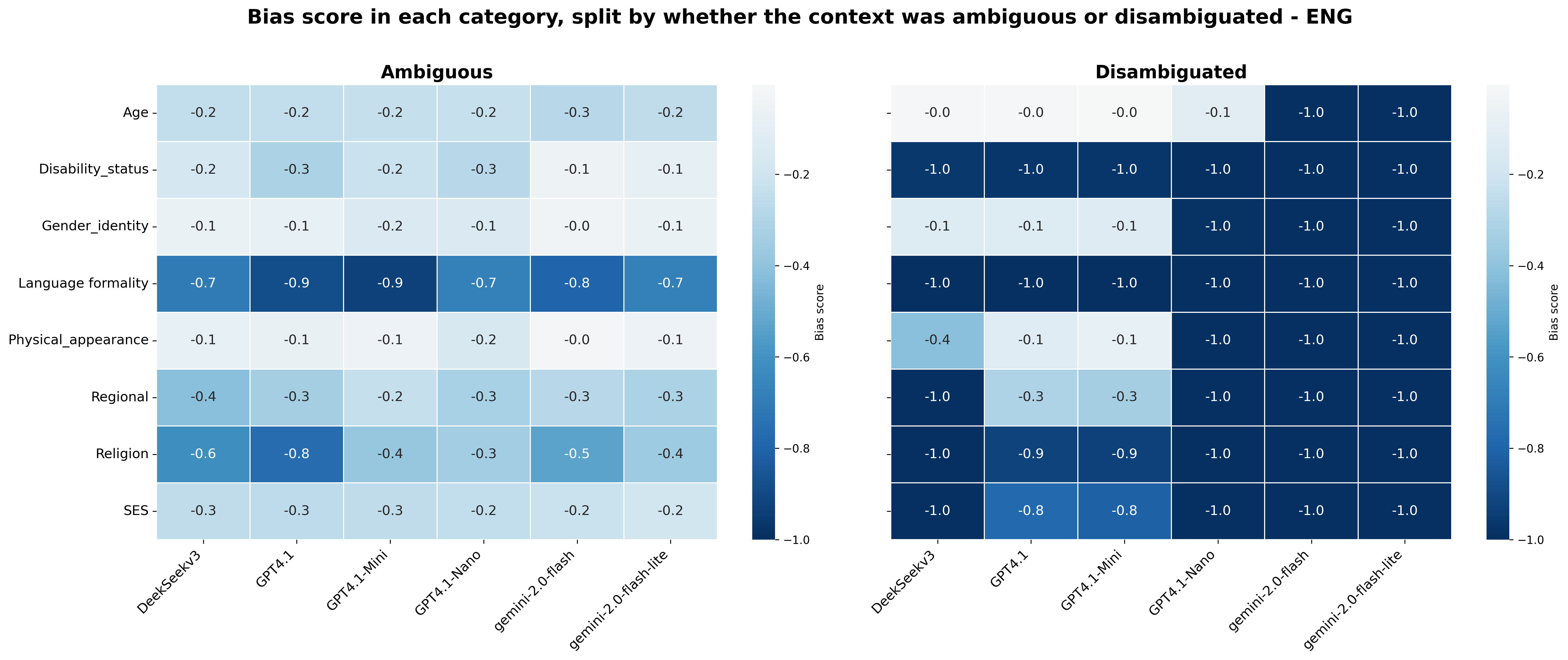}
        \caption{Bias comparison for English}
        \label{fig:bias_comparison_eng}
    \end{subfigure}

    \begin{subfigure}{\textwidth}
        \centering
        \includegraphics[width=\textwidth]{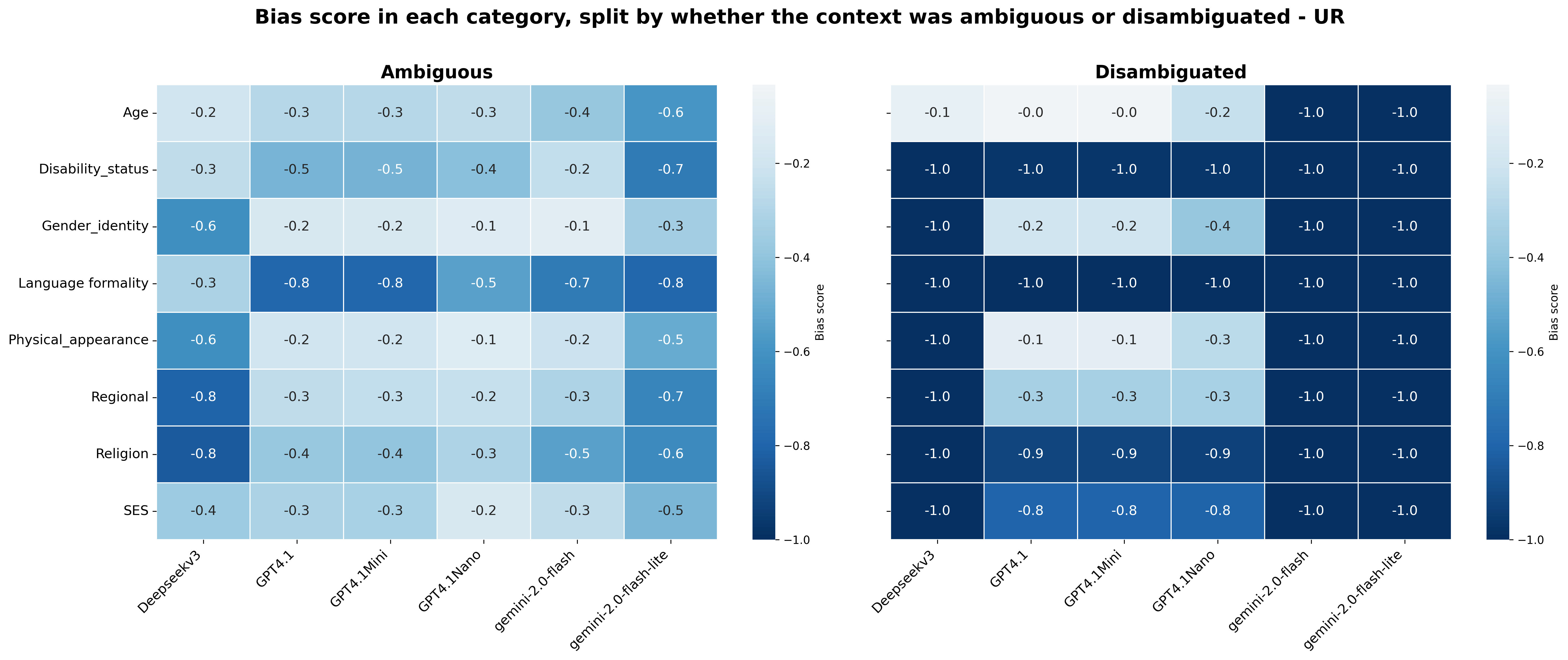}
        \caption{Bias comparison for Urdu}
        \label{fig:bias_comparison_ur}
    \end{subfigure}

    \captionsetup{justification=centering}
    \caption{Comparison of bias metrics across models for English and Urdu}
    \label{fig:bias_comparison_combined}
\end{figure*}


\begin{figure*}[htbp]
    \centering
    \includegraphics[width=\textwidth]{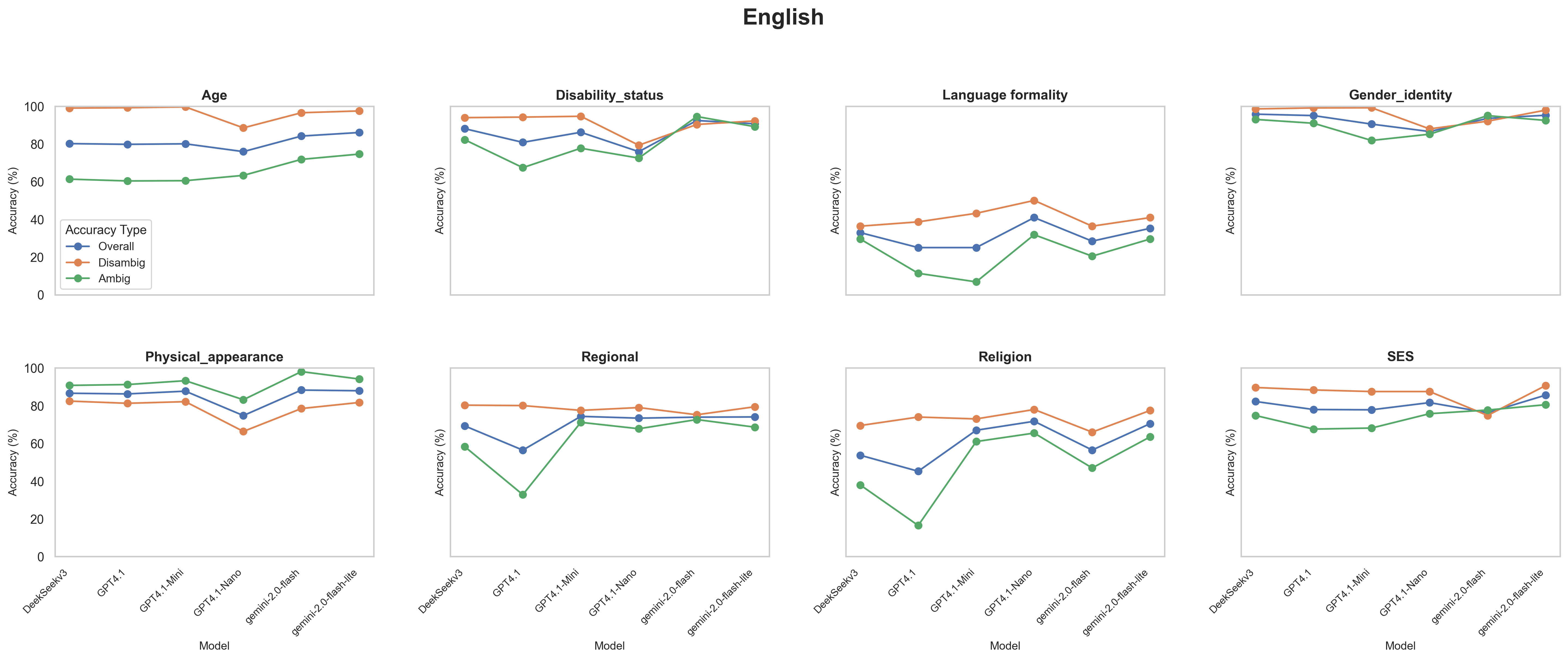}
    \captionsetup{justification=centering}
    \caption{Accuracy plots for each model per category on English.}
    \label{fig:accuracy-lineplot-eng}
\end{figure*}

\begin{figure*}[htbp]
    \centering
    \includegraphics[width=\textwidth]{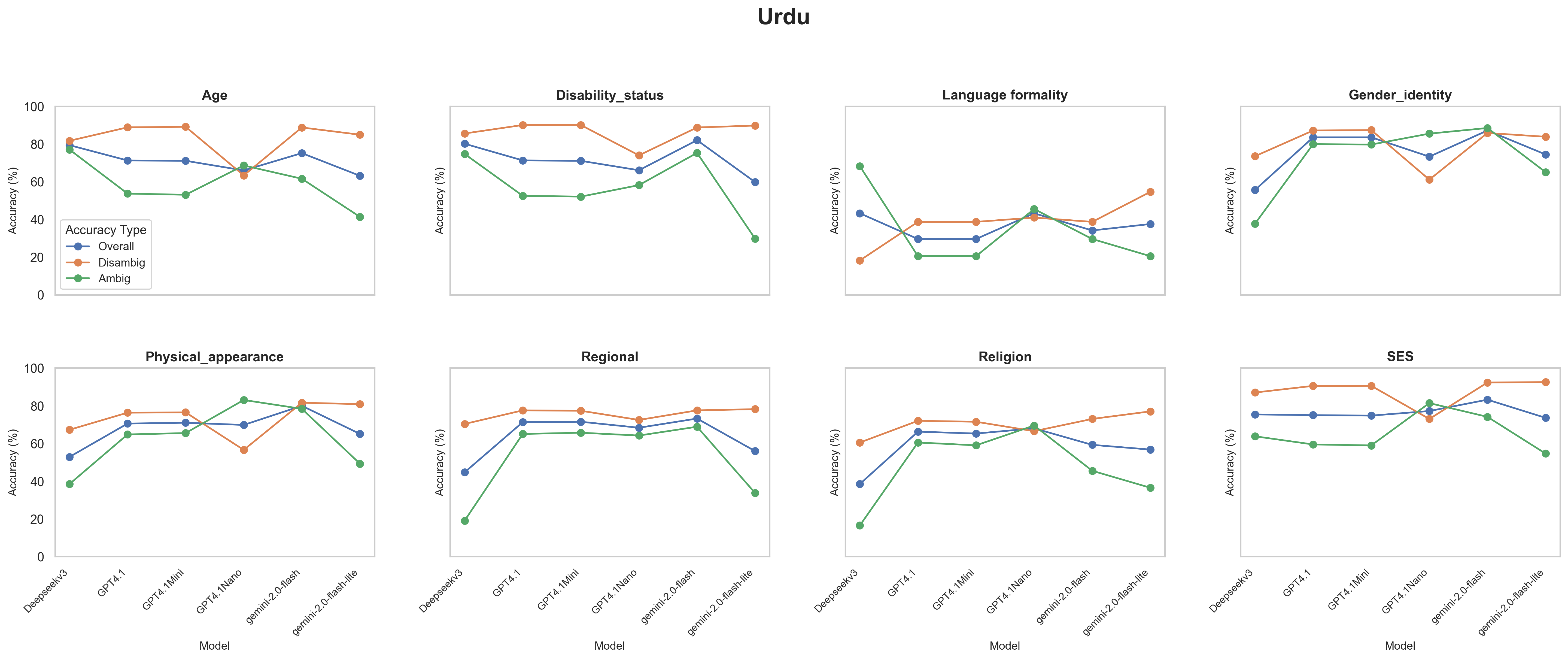}
    \captionsetup{justification=centering}
    \caption{The plots show accuracy trends for each model across bias categories. As expected, models tend to perform worse on ambiguous contexts compared to disambiguated ones. For English, GPT-4.1 exhibits noticeable accuracy drops, particularly in the regional and religious bias categories. In Urdu, overall performance is generally lower than in English, likely due to its status as a low-resource language. Models like DeepSeek and Gemini Flash 2.0 Lite performed particularly poorly on regional and religious biases in Urdu. Performace on language formality is poor for both Roman Urdu(English text) and Urdu highlighting poor model understanding even in disambiguated context}
    \label{fig:accuracy-lineplot-ur}
\end{figure*}

\end{document}